\documentclass{article}

\usepackage{microtype}
\usepackage{graphicx}
\usepackage{subfigure}
\usepackage{booktabs} 

\usepackage{hyperref}

\usepackage[accepted]{icml2020}

\begin{document}

\twocolumn[
\icmltitle{Our Evaluation Metric Needs an Update to Encourage Generalization}


\icmlsetsymbol{equal}{*}

\begin{icmlauthorlist}
\icmlauthor{Swaroop Mishra}{to}
\icmlauthor{Anjana Arunkumar}{to}
\icmlauthor{Chris Bryan}{to}
\icmlauthor{Chitta Baral}{to}
\end{icmlauthorlist}

\icmlaffiliation{to}{Department of Computer Science, Arizona State University}

\icmlcorrespondingauthor{Swaroop Mishra}{srmishr1@asu.edu}


\icmlkeywords{Machine Learning, ICML}

\vskip 0.3in
]



\printAffiliationsAndNotice 

\begin{abstract}
Models that surpass human performance on several popular benchmarks display significant degradation in performance on exposure to Out of Distribution (OOD) data. Recent research has shown that models overfit to spurious biases and `hack' datasets, in lieu of learning generalizable features like humans. In order to stop the inflation in model performance -- and thus overestimation in AI systems' capabilities -- we propose a simple and novel evaluation metric, \textit{WOOD Score}, that encourages generalization during evaluation.
\end{abstract}
\section{Introduction}
Training and evaluation in Machine Learning have been based on IID data. Encountering OOD data while testing is however, inevitable. This is due to several practical reasons such as: (i) difficulty in finding an `ideal' evaluation set that captures the entire distribution, and (ii) an `ideal' evaluation set identified during benchmark creation may not hold up against evolving benchmarks \cite{torralba2011unbiased, quionero2009dataset,  hendrycks2020pretrained}.

Recent language models have surpassed human performance across several popular AI benchmarks such as Imagenet \cite{russakovsky2015imagenet}, SNLI \cite{bowman2015large} and SQUAD \cite{rajpurkar2016squad}. However, a substantial performance drop is seen in these models on exposure to OOD data \cite{bras2020adversarial, eykholt2018robust, jia2017adversarial}. A growing number of recent works \cite{gururangan2018annotation, poliak2018hypothesis, kaushik2018much, tsuchiya2018performance, tan2019investigating,schwartz2017effect} have shown that models are not truly learning the underlying task; their performance in topping leaderboards can rather be attributed to their exploitation of spurious biases. This fundamental issue in learning leads to the overestimation of AI system performance \cite{bras2020adversarial, sakaguchi2019winogrande,hendrycks2019natural}, hence restricting AI deployment in several critical domains such as medicine \cite{hendrycks2016baseline}.

Several approaches have been proposed to address this issue at various levels: (i) \textit{Data} -- filtering of biases \cite{bras2020adversarial, li2019repair, li2018resound, wang2018dataset}, quantifying data quality, controlling data quality, using active learning, and avoiding the creation of low quality data \cite{mishra2020dqi, nie2019adversarial,gardner2020evaluating,kaushik2019learning}, and (ii) \textit{Model} -- utilizing prior knowledge of biases to train a naive model exploiting biases, and then subsequently training an ensemble of the naive model and a `robust' model to learn rich features \cite{clark2019don,he2019unlearn,mahabadi2019simple}. However, addressing this issue via an \textit{evaluation metric perspective} remains underexplored.

In pursuit of this metric, we first analyze recent works \cite{hendrycks2020pretrained, bras2020adversarial, wang2020train, hendrycks2019benchmarking, talmor2019multiqa} involving datasets that have been paired with an OOD counterpart, to isolate factors that distinguish OOD samples from IID samples. In this process, we identify four major problems that the Machine Learning community needs to address: ($q_{1}$) evaluating a model's generalization capability involves the \textit{overhead of OOD dataset identification}, ($q_{2}$) \textit{absence of a clear boundary separating IID from OOD} makes OOD identification even harder, ($q_{3}$) generalization is merely evaluated, not \textit{encouraged during evaluation}, and ($q_{4}$) the issue of \textit{inflated model performance} (and thus AI systems' overestimation) is yet to be addressed effectively.

Our analysis yields \textit{Semantic Textual Similarity (STS) of test data with respect to the training data as a distinguishing factor between IID and OOD}. We show the same across the prediction probabilities of ten models in two datasets. We then divide the datasets into several hierarchies, based on STS (and thus the degree of OOD characteristics). We can therefore reasonably identify samples within the dataset which have higher OOD characteristic levels. This allows the same dataset to be used to evaluate OOD ($q_{1}$). STS can not only be used to draw a boundary between IID and OOD, but also to control the degree of OOD characteristics in a dataset ($q_{2}$). We further propose a metric, \textit{Weighting Out of Distribution Score (WOOD Score)}, by weighting each test sample in proportion to its degree of OOD characteristics; higher levels of OOD characteristics imply higher weightage. Our intuition is simple: \textit{models must solve data with higher weightage to have higher accuracy}. This compels a model to generalize in order to dominate leaderboards ($q_{3}$). Our results show a decrease in model performance when evaluated using WOOD Score, resulting in lower benchmark accuracy ($q_{4}$). Our work inspires several potential solutions to handle OOD, using an evaluation metric perspective.



\section{Differentiating IID and OOD using STS}
\label{sec:diff}
We use two movie review datasets: SST-2 \cite{socher2013recursive} and IMDb \cite{maas2011learning}, which contain succinct expert reviews and full length general reviews respectively. We utilize SST-2 as the IID dataset and IMDb as the OOD dataset, and evaluate them using ten models: Bag-of-words (BoW) model \cite{harris1954distributional}, word embedding - word2vec \cite{mikolov2013distributed} and GloVe \cite{pennington2014glove} encoded with three models -word averages \cite{wieting2015towards}, LSTM \cite{hochreiter1997long} and CNN \cite{lecun1995convolutional}, and pretrained transformer models -BERT Base and Large \cite{devlin2018bert} along with GELU \cite{hendrycks2016gaussian} and RoBERTA \cite{liu2019roberta}, following a recent work on OOD Robustness \cite{hendrycks2020pretrained}. 

\vspace{-4mm}

\paragraph{Experimental Setup:} We perform a complete analysis by finding the STS between every pair of training set and test set samples. We sort samples of the test set in a descending order based on the average STS value with varying percentages of the train set samples. We consider the top 1\% -- 100\% of the training data (obtained by sorting train set samples in descending order of STS against each test set sample) with nine total steps, as similarity between the train and test sets is a task dependent hyperparameter, that trades off between inductive bias and spurious bias \cite{mishra2020dqi, gorman2019we}. We train models on the IID data (SST-2) and evaluate on both the IID test set (SST-2) and the OOD test set (IMDb). We compare model predictions with the average STS value for each sample.

\begin{figure}[H]
    \centering
    \includegraphics[width=\columnwidth]{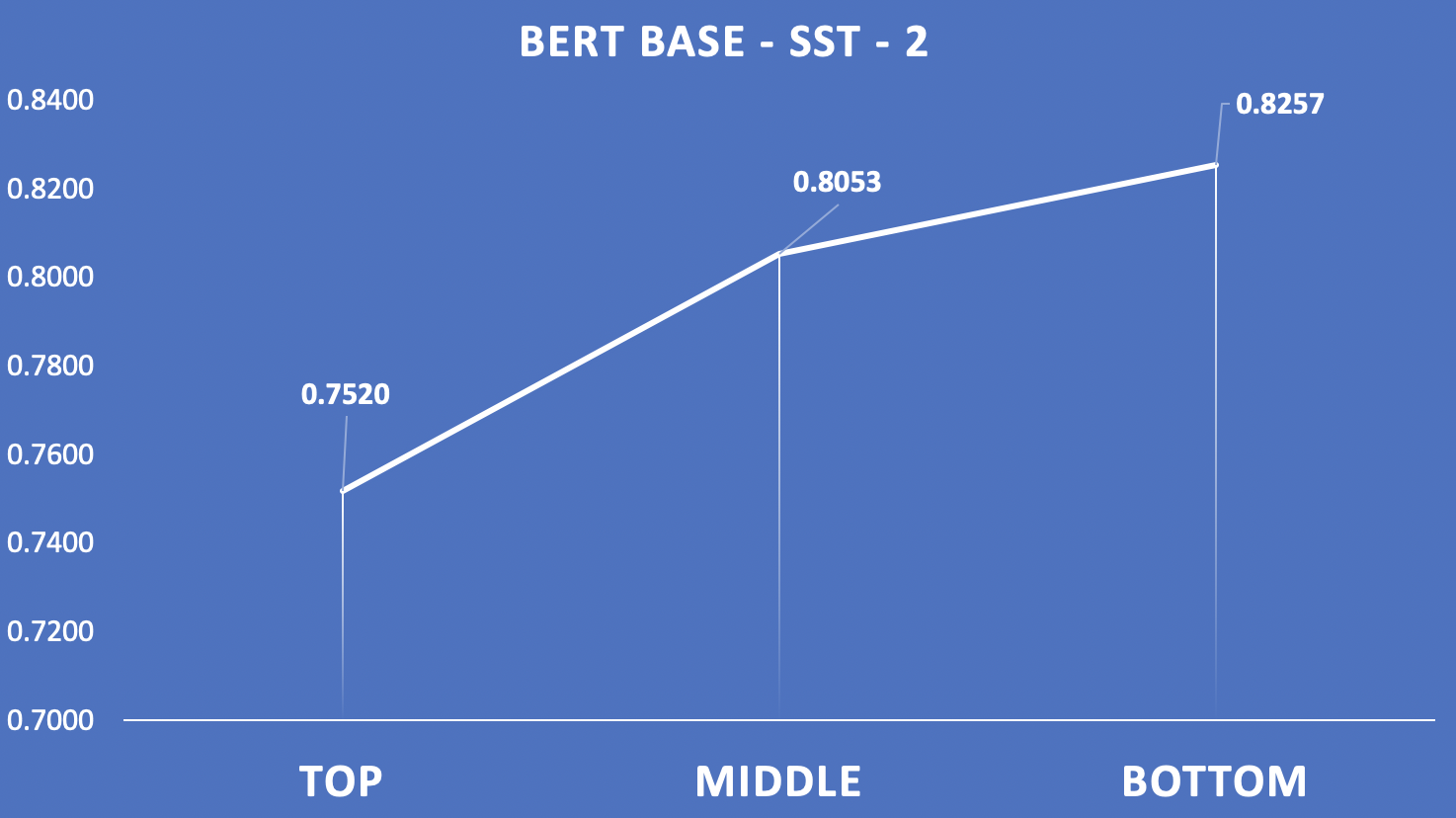}
\vspace{-4mm}
    \caption{Softmax probabilities of incorrect classifications using BERT-Base model across test samples of SST-2 and IMDB in decreasing order train (SST-2)-test similarity.}
    \label{fig:15}
\vspace{-4mm}
\end{figure}

\begin{figure}[H]
    \centering
    \includegraphics[width=\columnwidth]{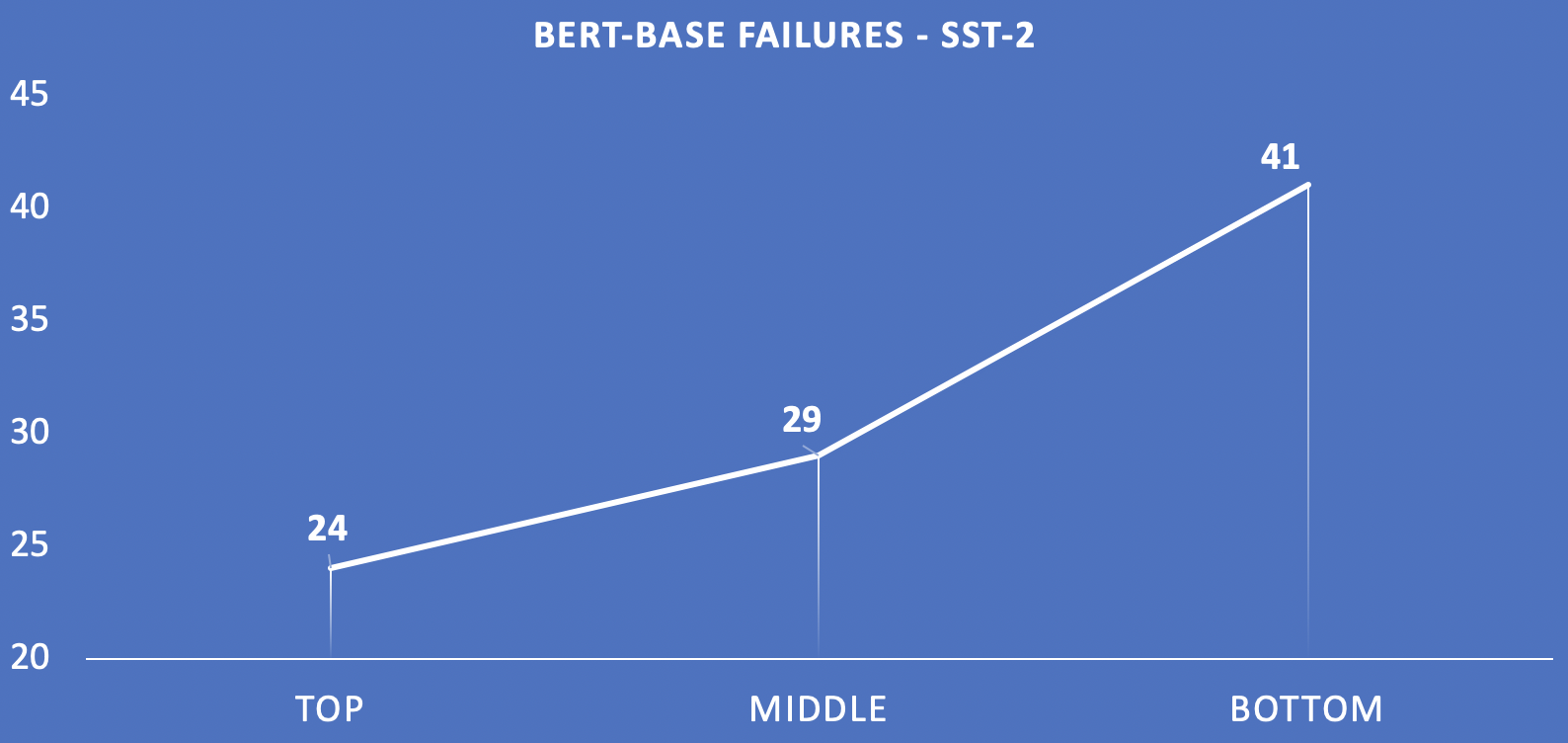}
    \includegraphics[width=\columnwidth]{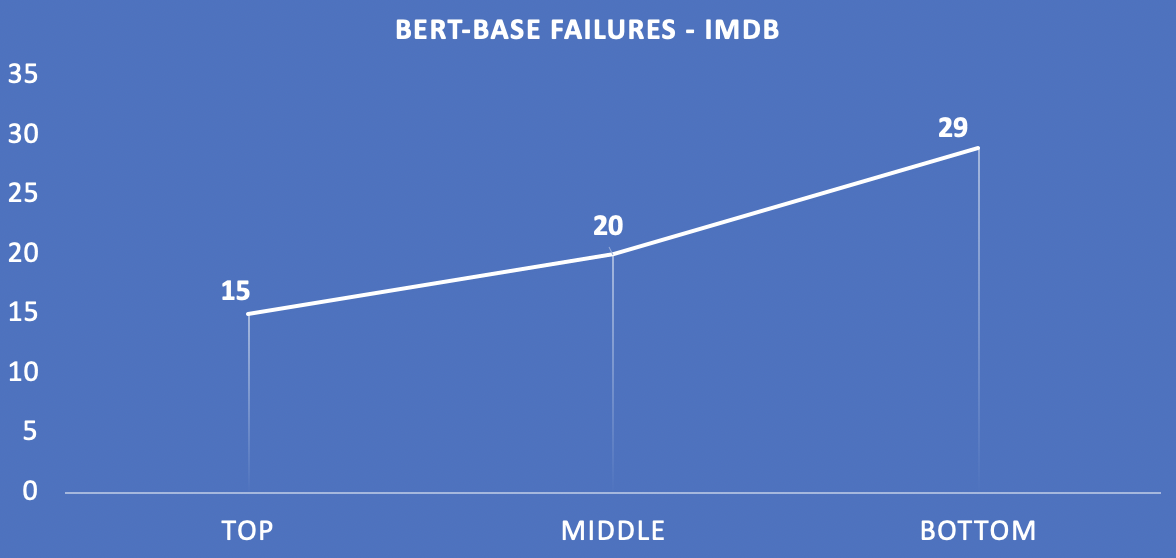}
\vspace{-4mm}
    \caption{Number of incorrect classifications using BERT-Base model across test samples of SST-2 and IMDB in decreasing order of train (SST-2)-test similarity.}
    \label{fig:2}
\vspace{-4mm}
\end{figure}

\begin{figure}[H]
    \centering
    \includegraphics[width=\columnwidth]{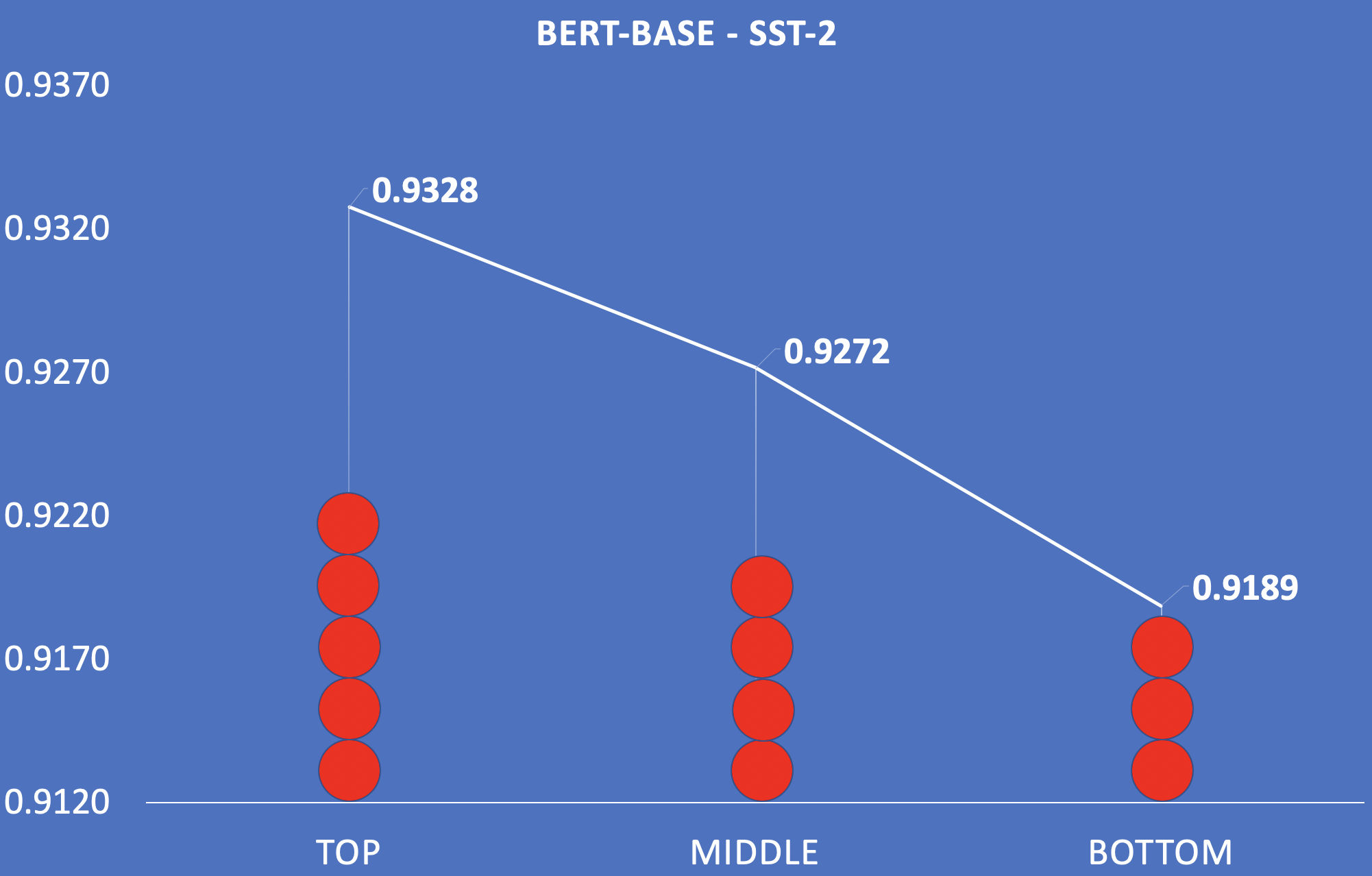}
    \includegraphics[width=\columnwidth]{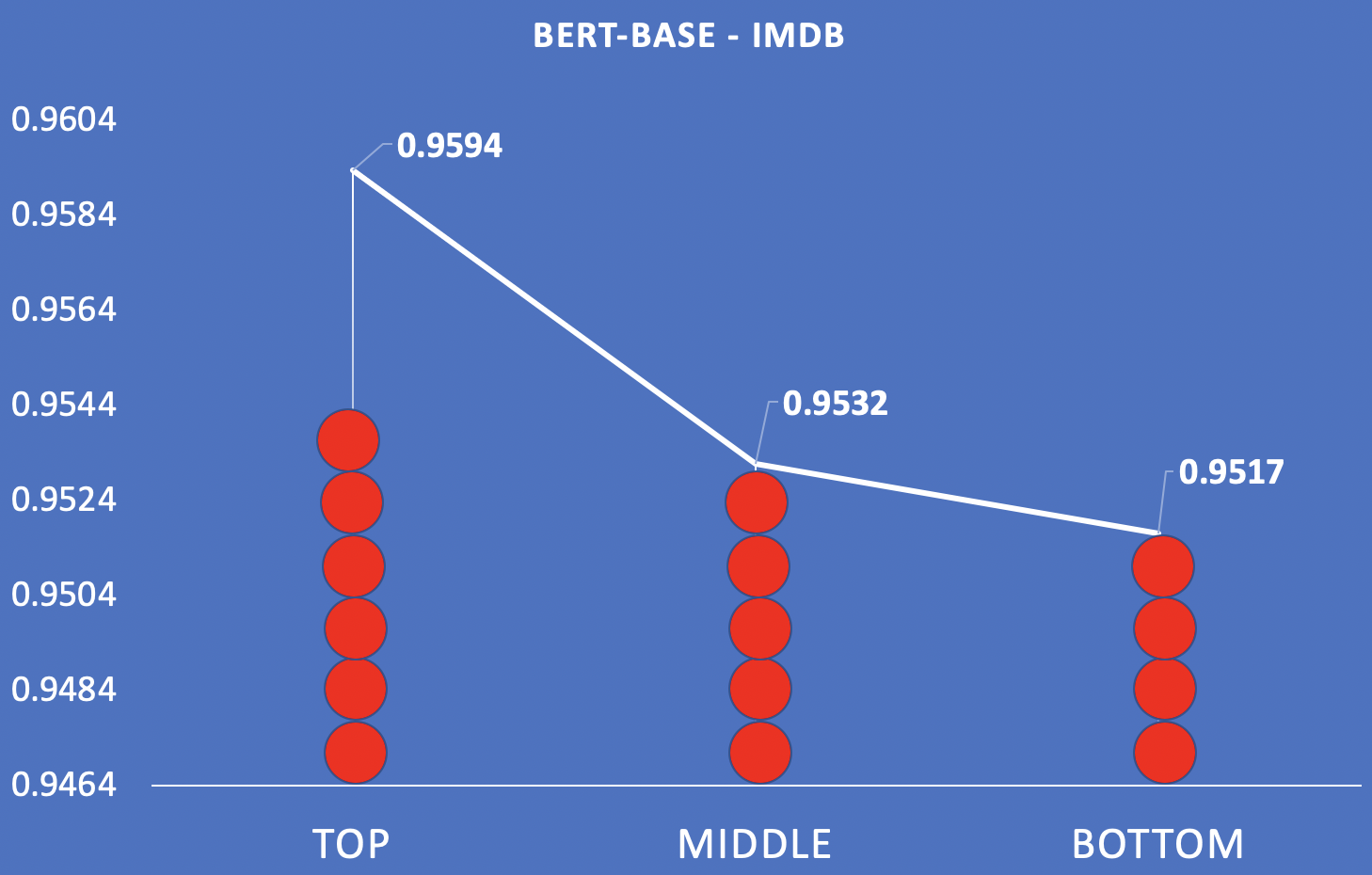}
    \vspace{-5mm}
    \caption{Softmax probabilities of correct classifications using BERT-Base model across test samples of SST-2 and IMDB in decreasing order of train (SST-2)-test similarity. The circles help visualize the impact of uncertainty in classification by representing the potential number of incorrectly classified samples per 20 samples for each split of data.}
    \label{fig:3}
\vspace{-8mm}
\end{figure}

\begin{figure*}
    \includegraphics[width=2\columnwidth]{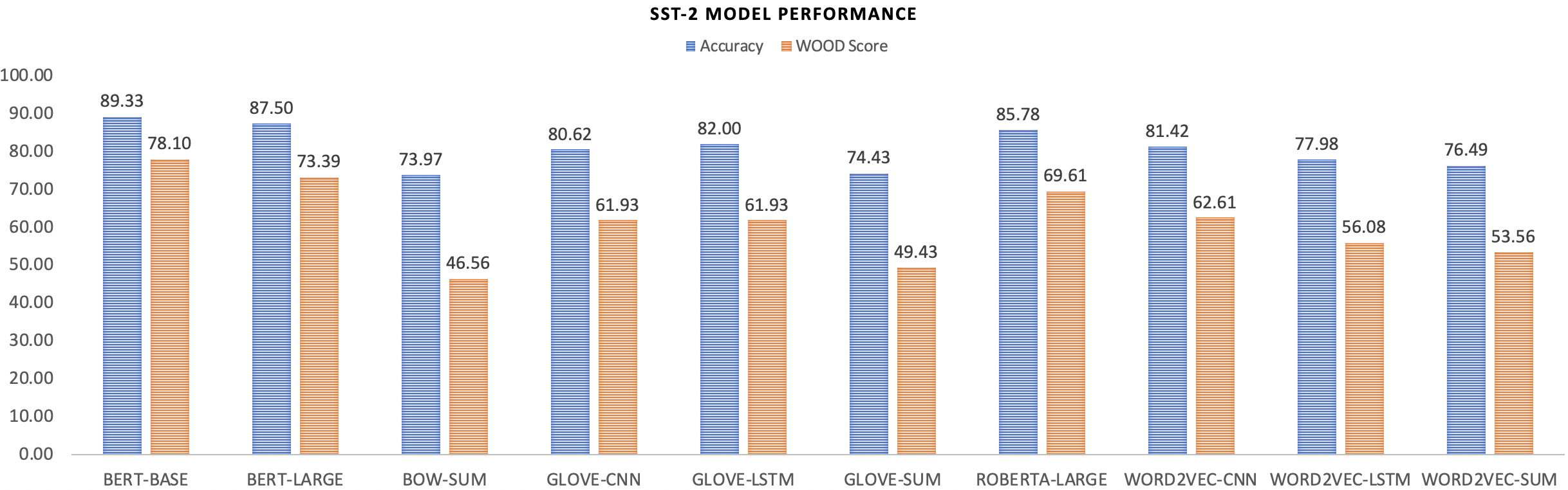}
    \includegraphics[width=2\columnwidth]{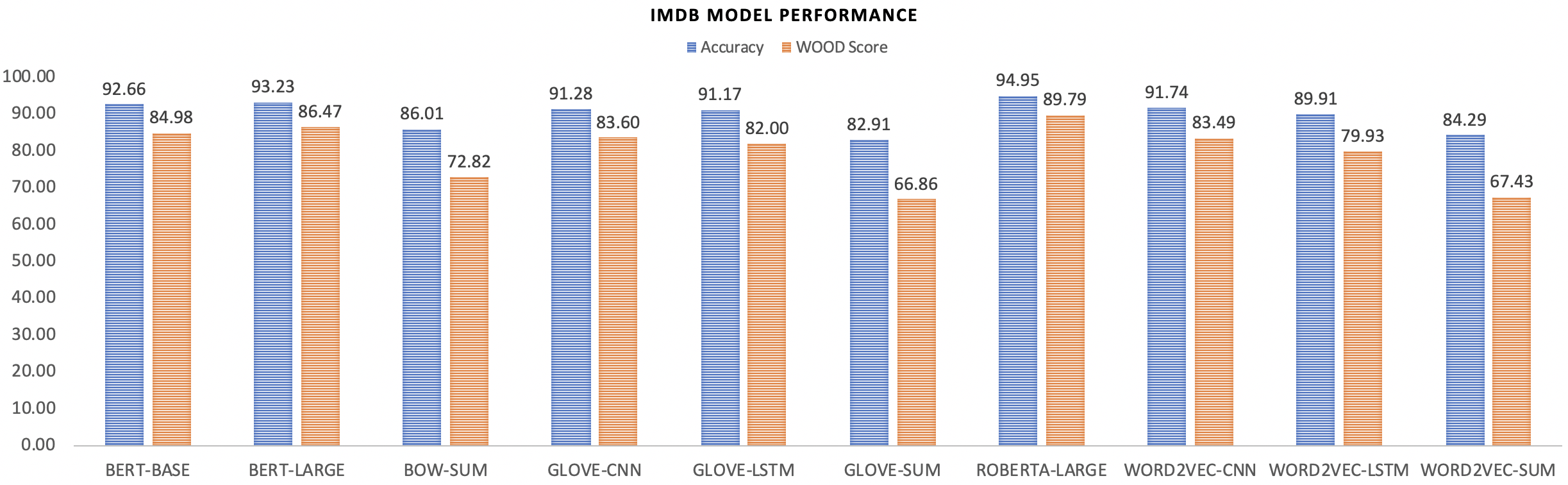}
    \caption{Accuracy and WOOD Score of SST-2 and IMDB across models.}
    \label{fig:4}
\vspace{-4mm}
\end{figure*}

\paragraph{Results:} We find three broad patterns: (i) Samples with higher average STS value are classified correctly with higher confidence (Figure \ref{fig:3}), (ii) Incorrect classification frequency increases as STS value decreases  (Figure \ref{fig:2}), and (iii) Confidence value (softmax probability) of an incorrect classification increases as we move towards the region having lower STS  (Figure \ref{fig:15}).

\section{Metric for Equitable Evaluation of Data} 
Benchmarks and leaderboards guide model development. We propose \textit{equitable data evaluation} using a weighted metric, in lieu of the conventional metric that does uniform evaluation. Our intuition is to weight the evaluation score of a sample in proportion to its `hardness', i.e. the level of OOD characteristics it displays. In this paper, we define a metric, \textit{WOOD Score},  using STS (with the training set) as an inverse indicator of `hardness', based on the findings of Section \ref{sec:diff}.

\paragraph{Formalization:}
Let $X$ represent a dataset where $X_{Test}$ is the test set spanned by $i$ and $X_{Train}$ is the train set. $E$ represents the evaluation metric (which depends on the application), $p$ is the degree of OOD characteristics a sample has, $S$ represents STS, $a$ allows for the control of $p$ based on $S$, $b$ is the number of train samples considered that have higher similarity values than rest of the dataset. $W_{opt}$ represents our proposed metric in generic form, and $W_{acc}$ is the proposed accuracy metric in this paper. $A$ represents accuracy. $A_{3}$, $A_{2}$, and $A_{1}$ represent the accuracies of the categories of samples having the highest, moderate, and lowest degrees of OOD characteristics respectively.

\begin{equation}
W_{opt}=\sum_{X_{Test}} E_{i}p_{i}
\end{equation}
\begin{equation}
p=\frac{a}{\sum_{X_{Train}}{\max\limits_{b}S}}
\end{equation}
\begin{equation}
W_{acc}=A_{1}+2A_{2}+3A_{3}
\end{equation}
\vspace{-1mm}

\paragraph{Controlling Benchmark Accuracy Using Hyperparameters:} Benchmark accuracy can be controlled using appropriate values of $a$ and $b$. Using $W_{acc}$ for both datasets across ten models has resulted in a significant reduction in accuracy, as illustrated in Figure \ref{fig:4}.

\section{Discussion}
\paragraph{Extension to Other Metrics Beyond Accuracy: } Our focus in this paper is on accuracy, but the same issue of `uniform evaluation score' prevails in all metrics such as Pearson's correlation score, BLEU, and $F_{1}$ score. A potential future work is to extend our metric beyond accuracy.


\begin{figure}[H]
    \centering
    \includegraphics[width=\columnwidth]{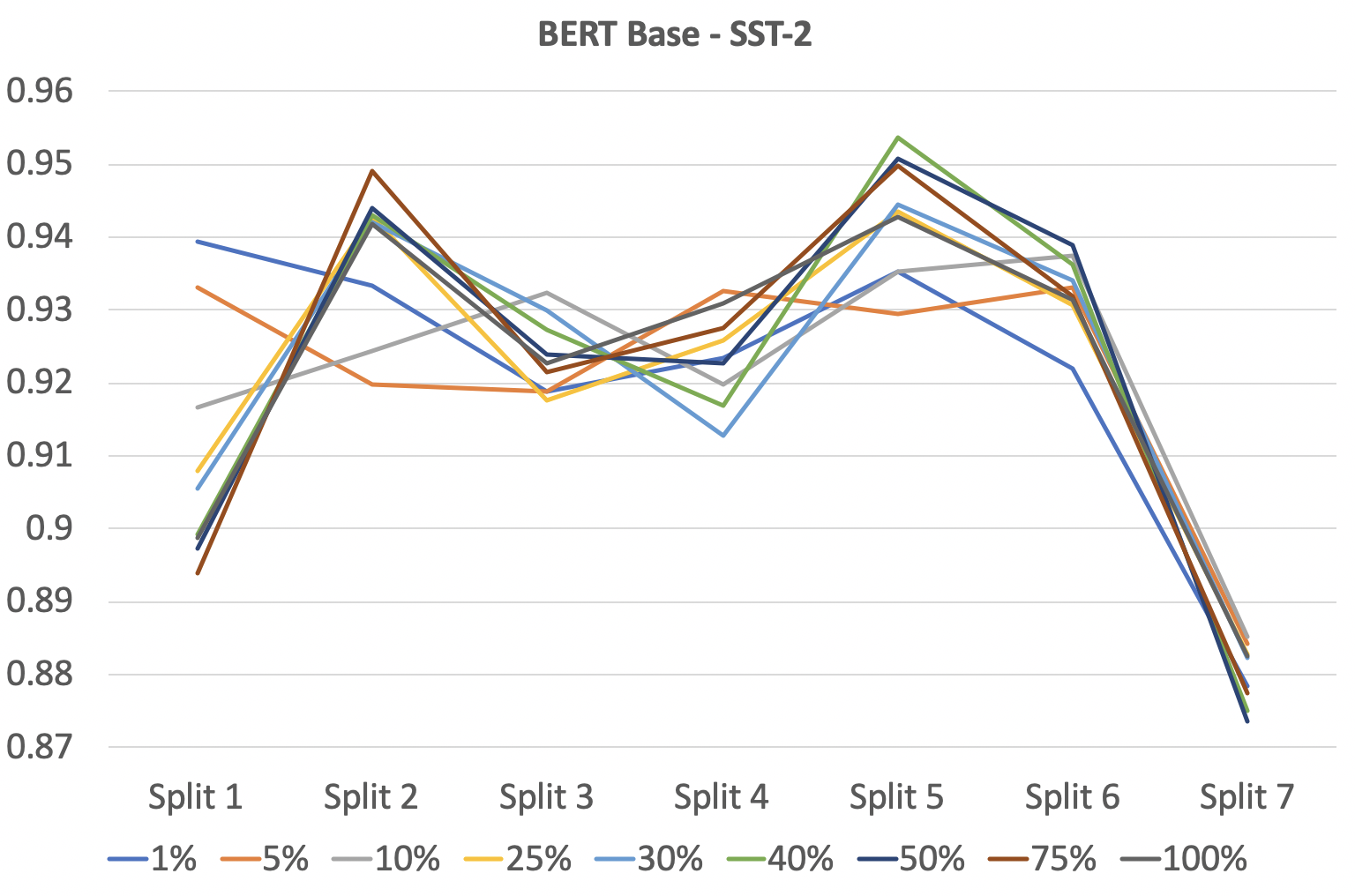}
    \includegraphics[width=\columnwidth]{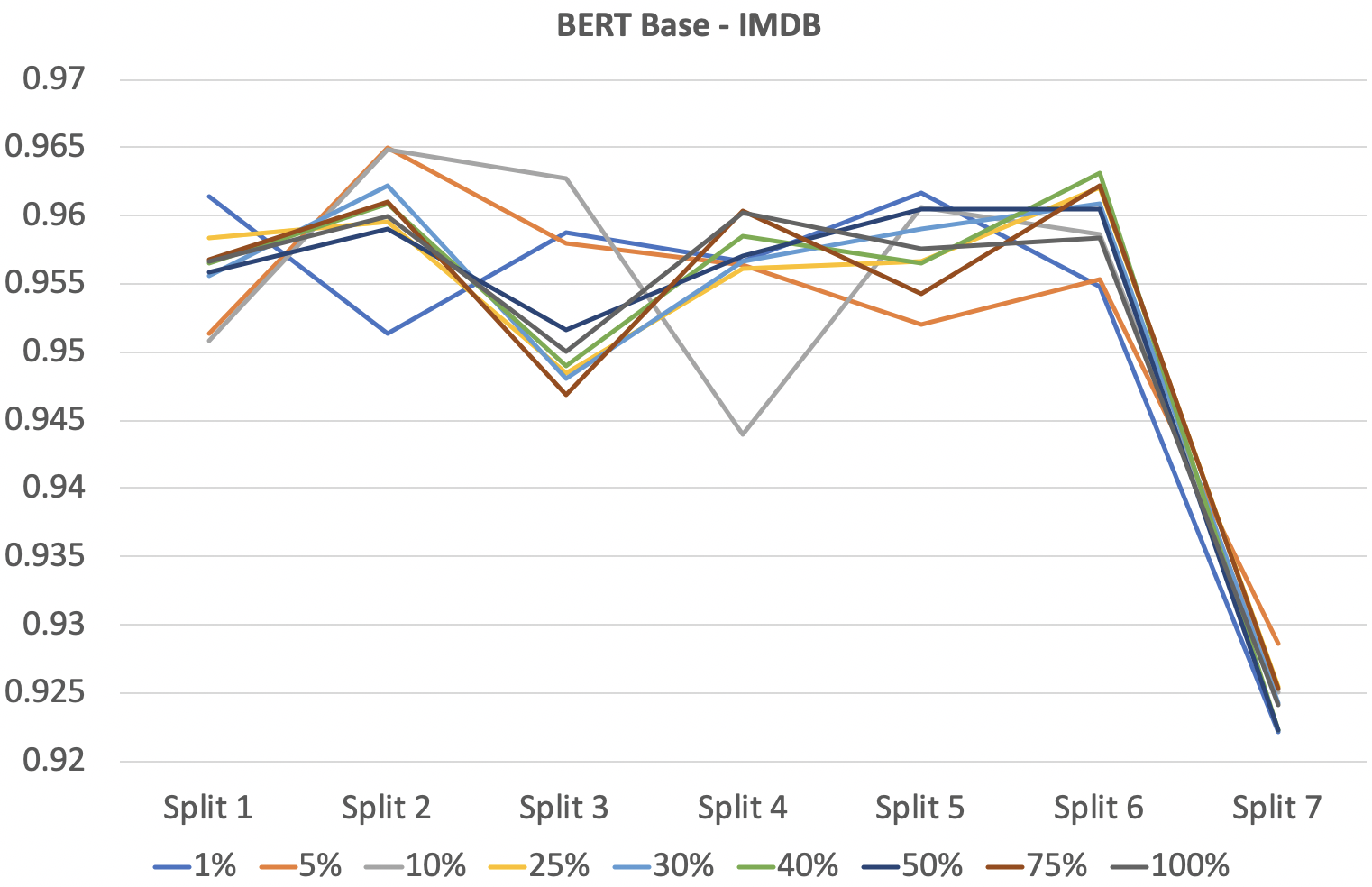}
\vspace{-4mm}
    \caption{The top $b$\% of training samples is obtained by sorting in descending order of STS with each test set sample. The test set samples are then divided into seven splits, based on decreasing STS values averaged over the top $b$\% of training samples considered.}
    \label{fig:5}
\vspace{-4mm}
\end{figure}

\begin{figure}[H]
    \centering
    \includegraphics[width=\columnwidth]{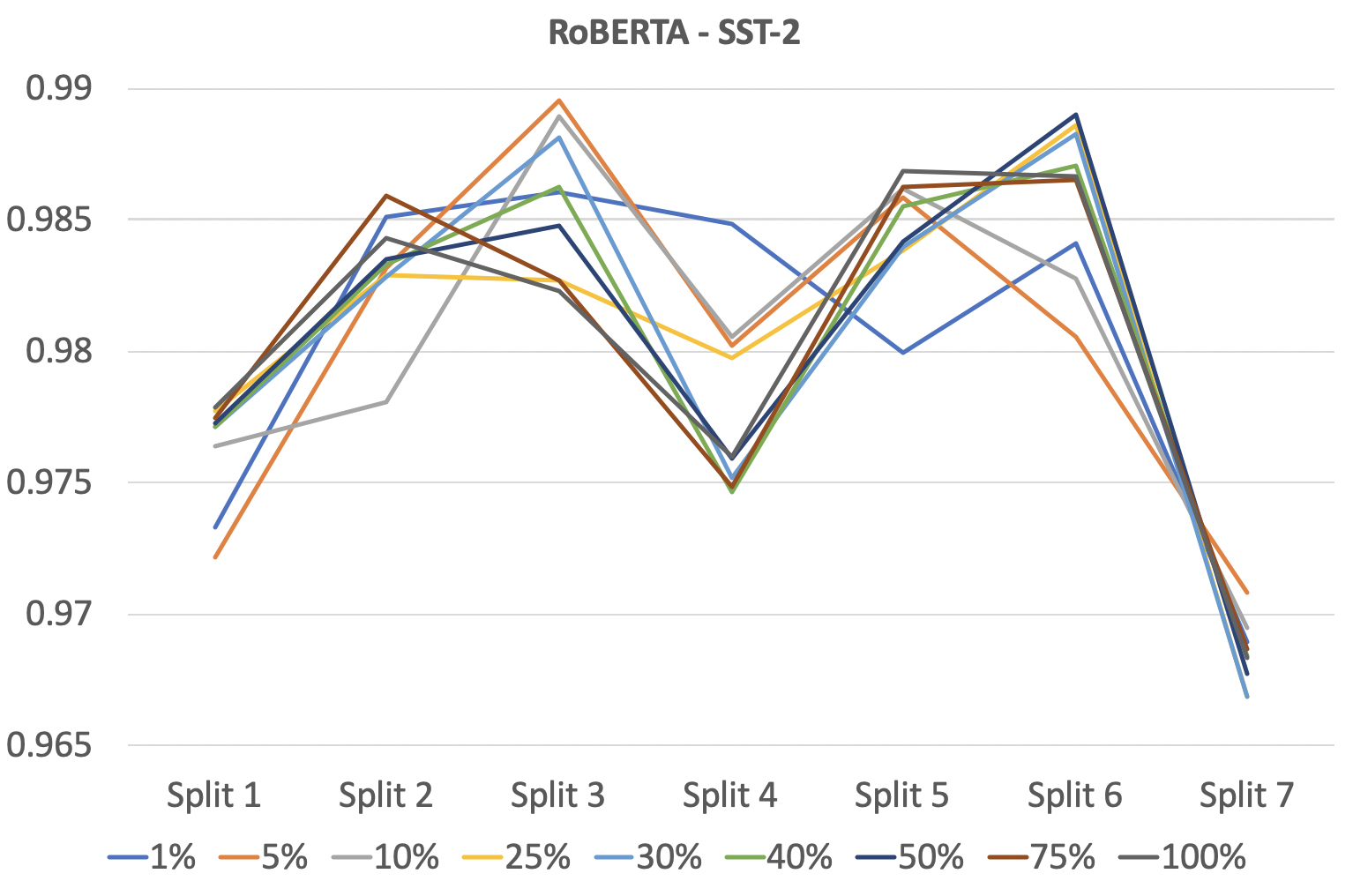}
    \includegraphics[width=\columnwidth]{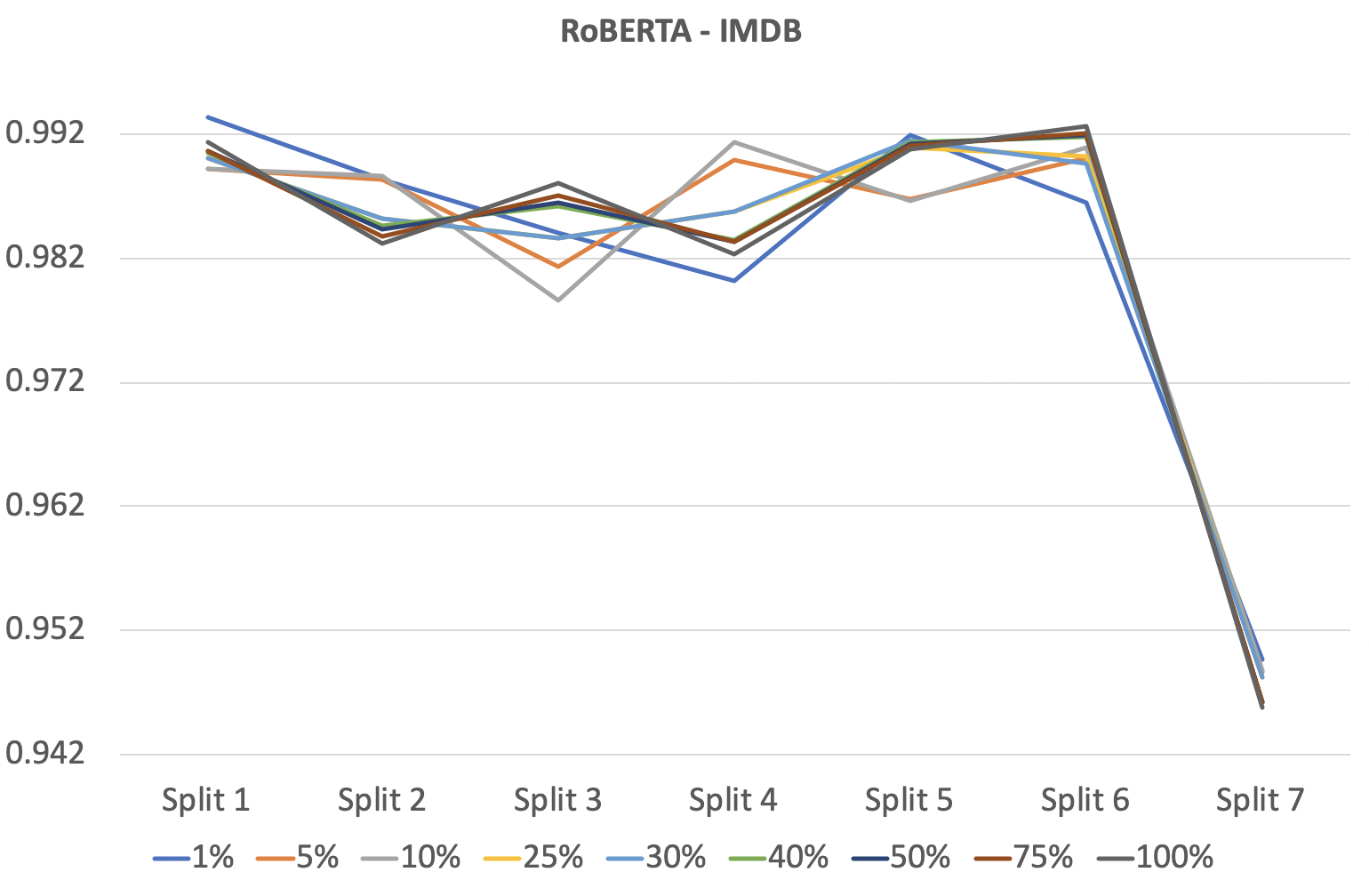}
\vspace{-6mm}
    \caption{Detailed analysis where the test set is divided in to seven splits. Figure shows variation across a range of $b$ values for the RoBERTA model.}
    \label{fig:6}
\vspace{-5mm}
\end{figure}

\textbf{Augumenting STS:} Detailed analysis shows that STS's distinguishing capacity may not follow monotonic behavior for certain cases, such as those illustrated in Figure \ref{fig:5} and \ref{fig:6}. Similarity across several granularities -- such as word, bigram, and trigram -- can be used to augument STS and increase the robustness of `hardness' evaluation.\\
\textbf{OOD Drop is More Than IID with the Updated Metric: }
Figure \ref{fig:5} illustrates that Benchmark accuracy drops more in the IMDb dataset (OOD) than SST-2 (IID). IMDb has a different writing style and sample size in contrast to SST-2. This may indicate that, our metric is more effective for applications involving diverse data considered as OOD.\\
\textbf{Leveraging the Metric to Learn a Task Better: } Hardness of an anticipated OOD task can be reduced by adding similar data to the training set. Our metric can guide the data augmentation process to learn a task better.\\
\textbf{Strengthening `in-house' IID (acting OOD): }
We further observe that, IID data, even with STS calibration, may not represent many properties of an OOD data sample -- such as variations in writing style, topic, vocabulary , sentence length, and number of sentences. Contrast sets \cite{gardner2020evaluating} can be utilized to partially address this issue by strengthening the test set. We recommend that dataset creators go beyond the common patterns found in a dataset, and draw patterns from other

 datasets intended for the same task, while creating contrast sets, in order to fully address this issue. Robustness of a dataset can also be improved by adding more application-centric variations -- such as anticipating various possibilities in question format, structure, language, reasoning skills required, syntax, and numerical reasoning requirements.

\vspace{-4mm}
\section{Conclusion}
We propose a STS based approach to address certain important issues in robustness research: finding a suitable OOD dataset, and drawing a valid boundary betweeen IID and OOD. Our approach also helps in controlling the degree of OOD characteristics. We also propose \textit{WOOD Score}, a metric that implicitly encourages generalization by weighting the evaluation scores of samples in proportion to their hardness. This results in reduced model performance and benchmark accuracy, addressing the issue of model performance inflation and overestimation of progress in AI. We show the efficacy of our work on ten popular models across two NLP datasets. Finally, we provide insights into several future works on encouraging generalization and improving robustness from a metric perspective. 

\nocite{langley00}
\section*{Acknowledgements}

We thank the anonymous reviewers, Dan Hendrycks (UC Berkeley) and Xiaoyuan Liu (Shanghai Jiao Tong University) for their thoughtful feedback. We also thank Jason Yalim and ASU HPC for their consistent support. The support of DARPA SAIL-ON program (W911NF2020006)  is gratefully acknowledged.

\bibliography{example_paper}
\bibliographystyle{icml2020}

\end{document}